\documentclass[conference]{IEEEtran}
\IEEEoverridecommandlockouts    % to override locked commands

\usepackage{multirow}
\usepackage{siunitx}
\usepackage{tabularx} % for 'tabularx' environment
\usepackage{makecell} % text formatting within individual table cells
\usepackage{booktabs}
\usepackage{fp}     % fixed point arithmetic for scaling
\usepackage{pifont}% http://ctan.org/pkg/pifont

\usepackage{subcaption}

\usepackage{hyphenat}% \nohyphens
\usepackage[table]{xcolor}
\usepackage[nolist,nohyperlinks,printonlyused]{acronym}

\usepackage[export]{adjustbox}
\usepackage{esvect}

\usepackage{svg}

\usepackage{graphicx}
\usepackage{amsmath}
\usepackage{amssymb}
\usepackage{booktabs}
\usepackage{collcell}
\usepackage{pgf}
\usepackage{xparse}
\pgfkeys{/pgf/number format/.cd,fixed,fixed zerofill,precision=2}
\newcolumntype{E}{>{\collectcell\usermacro}r<{\endcollectcell}}
\newcommand\usermacro[1]{\pgfmathparse{100*#1}\pgfmathprintnumber\pgfmathresult}

\makeatletter
\@namedef{ver@everyshi.sty}{}
\makeatother

\usepackage{nicematrix}
\usepackage{tikz}
\usetikzlibrary{positioning,shadows,intersections,fit}

\usepackage{graphbox}  % used for align keyword in includegraphics for fig 3

\usepackage[framemethod=tikz]{mdframed} % paragraph highlight boxes

\usepackage{soul}	% text highlight
\usepackage[normalem]{ulem} % strikethrough
\usepackage{etoolbox}  % used for toggling between two versions
\newtoggle{draft_mode}
\toggletrue{draft_mode}

\iftoggle{draft_mode}
{
    \usepackage{todonotes}
}{
    \usepackage[disable]{todonotes}
}

\usepackage[pagebackref,breaklinks,colorlinks,hidelinks]{hyperref}
\usepackage{orcidlink}  % to scale orcid icon

\usepackage[capitalize]{cleveref}
\crefname{section}{Sec.}{Secs.}
\Crefname{section}{Section}{Sections}
\Crefname{table}{Table}{Tables}
\crefname{table}{Tab.}{Tabs.}

\usepackage{mwe} % for example-image
\newcommand{\kitti}{\emph{KITTI}}

\newcommand{\nuscenes}{\emph{nuScenes}}

\newcommand{\dkitti}{KITTI}
\newcommand{\dnuscenes}{nuScenes}
\newcommand{\dwaymo}{Waymo}

\newcommand{\sensortosensor}{\textit{sensor-to-sensor}}
\newcommand{\datasettodataset}{\textit{dataset-to-dataset}}

\newcolumntype{R}[2]{%
    >{\adjustbox{angle=#1,lap=\width-(#2)}\bgroup}%
    l%
    <{\egroup}%
}
\definecolor{car}{rgb}{0.39215686274509803, 0.5882352941176471, 0.9607843137254902}
\definecolor{bicycle}{rgb}{0.39215686274509803, 0.9019607843137255, 0.9607843137254902}
\definecolor{motorcycle}{rgb}{0.11764705882352941, 0.23529411764705882, 0.5882352941176471}
\definecolor{truck}{rgb}{0.3137254901960784, 0.11764705882352941, 0.7058823529411765}
\definecolor{othervehicle}{rgb}{0.0, 0.0, 1.0}
\definecolor{person}{rgb}{1.0, 0.11764705882352941, 0.11764705882352941}
\definecolor{bicyclist}{rgb}{1.0, 0.1568627450980392, 0.7843137254901961}
\definecolor{motorcyclist}{rgb}{0.5882352941176471, 0.11764705882352941, 0.35294117647058826}
\definecolor{road}{rgb}{1.0, 0.0, 1.0}
\definecolor{parking}{rgb}{1.0, 0.5882352941176471, 1.0}
\definecolor{sidewalk}{rgb}{0.29411764705882354, 0.0, 0.29411764705882354}
\definecolor{otherground}{rgb}{0.6862745098039216, 0.0, 0.29411764705882354}
\definecolor{building}{rgb}{1.0, 0.7843137254901961, 0.0}
\definecolor{fence}{rgb}{1.0, 0.47058823529411764, 0.19607843137254902}
\definecolor{vegetation}{rgb}{0.0, 0.6862745098039216, 0.0}
\definecolor{trunk}{rgb}{0.5294117647058824, 0.23529411764705882, 0.0}
\definecolor{terrain}{rgb}{0.5882352941176471, 0.9411764705882353, 0.3137254901960784}
\definecolor{pole}{rgb}{1.0, 0.9411764705882353, 0.5882352941176471}
\definecolor{trafficsign}{rgb}{1.0, 0.0, 0.0}

\newcommand{\cD}{\mathcal{D}}

\newcommand{\figref}[1]{Fig.~\ref{#1}}    % already defined

\newcommand{\tabref}[1]{Table~\ref{#1}}    % already defined

\makeatletter
\def\onedot{.}
\def\etal{et~al\onedot}

\makeatother

\definecolor{darkgreen}{rgb}{0,0.7,0}

\usepackage{multirow}
\usepackage{lipsum}
\usepackage{amsmath}
\usepackage{amssymb}
\usepackage[ruled,vlined]{algorithm2e}
\usepackage{graphicx}

\title{\LARGE \bf Robust \acs{lidar} Sensor-to-Sensor Domain Generalization via Accumulated-Point Scene Completion}

\author{
	\parbox{\textwidth}{%
		\centering
 Christoph B. Rist$^{1}$\orcidlink{0000-0003-4578-1616}, %
 Larissa T. Triess$^{}$ \orcidlink{0000-0003-0037-8460}, %
 Markus Enzweiler$^{2}$ \orcidlink{0000-0001-9211-9882}, %
 Dariu M. Gavrila$^{1}$ \orcidlink{0000-0002-1810-4196}%
	}%
	\thanks{$^{1}$Technical University Delft, Intelligent Vehicles Group, The Netherlands}%
	\thanks{$^{2}$Hochschule Esslingen, Esslingen, Germany}%
}

\begin{document}

\begin{acronym}
\acro{da}[DA]{domain adaptation}
\acro{dg}[DG]{domain generalization}
\acro{uda}[UDA]{unsupervised domain adaptation}
\acro{lidar}[LIDAR]{light detection and ranging}
\end{acronym}

%%%%%%%%% TITLE
\maketitle
\thispagestyle{empty}
\pagestyle{empty}

\maketitle
% !TEX root = ../root.tex
%%

\begin{abstract}
We study domain generalization for \acs{lidar} semantic segmentation.
Deep neural networks achieve strong performance on 3D point clouds, yet real-world \acs{lidar} exhibits sensor-specific sampling patterns that hinder transfer across datasets.
In many applications, performance on completely unseen sensors is crucial.
A model should preferably work on a new setup without prior access to its data or labels.
To this end, we introduce a source-only training scheme that adds an occupancy prediction auxiliary objective.
We aggregate source-domain scans into a simple world model and use it as the supervision signal, replacing single-scan standard segmentation training.

A key property of our approach is that it does not alter the input scan.
During training and testing the network consumes the same single-frame \acs{lidar} input, only the supervision signal is enriched via the occupancy objective.
At deployment, there is no runtime overhead as the auxiliary head can be dropped for pure segmentation, or kept to obtain scene completion alongside semantic segmentation.
In an experimental evaluation across distinct \acs{lidar} datasets, we demonstrate that a model's generalization to new domains can be significantly affected by how it is supervised.
In certain setups, performance on unseen domains improves without any explicit \acl{da} (no target scans, no target labels, no sensor priors) and without prediction refinement.
At the same time, our method maintains competitive source-domain segmentation performance and incurs no additional inference cost.
\end{abstract}

% !TEX root = ../root.tex
%%

%------------------------------------------------------------------------
\section{Introduction}

%------------------------------------------------------------------------
% What is the problem?
Highly automated vehicles and mobile robots rely on deep learning–based perception models that typically require large volumes of annotated data.
Yet these systems must operate across diverse domains induced by weather, geography, traffic culture, and sensor generations.
Since data collection and annotation is costly and time-consuming, repeating them for every new domain is impractical.
\emph{\Ac{uda}} targets this challenge with automated methods that adapt models to new domains without requiring manual annotations in the target domain.
Going further, \ac{dg} seeks to build models using only source-domain data without any access to potential target domains (neither labels nor raw sensor data) and to maximize performance on these unseen targets at test time.
Models that generalize well to unseen domains reduce retraining efforts when experimenting with new platforms and sensor setups.
They accelerate deployment and iteration, and they improve robustness in safety critical systems.

% !TEX root = ../root.tex
%%
\begin{figure}[t]
  \centering
  % Use the base name without .svg; width spans both columns here
  \includegraphics[width=\linewidth]{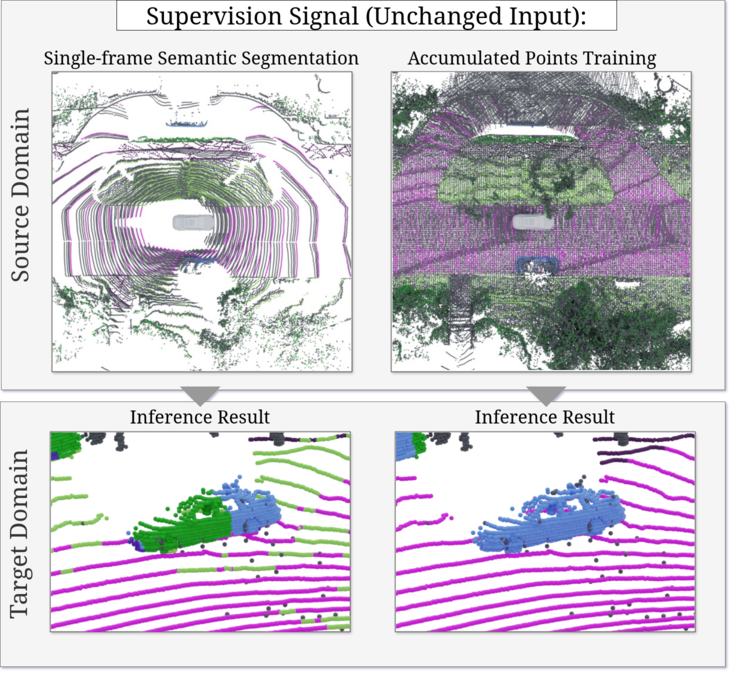} % 
  \caption{%
  \textbf{Sensor-to-sensor Domain Generalization with richer supervision:} %
  Comparing two training signals on the source domain (\dnuscenes{}) and their effect on the target domain (\dwaymo{}) segmentation task (blue: car, green: vegetation, magenta: ground).
  The left column uses standard per-scan semantic segmentation, the right column uses accumulated multi-scan points as supervision signal.
  When evaluated on an unseen target domain sensor (bottom row), the accumulated points training is robust against sensor ground patterns and yields correct predictions with consistent boundaries.
  }
  \label{fig:title}%
\end{figure}

Generic, feature-based \ac{da} techniques struggle with \acs{lidar} domain shifts dominated by sampling-pattern changes, motion distortion, and mounting geometry~\cite{Triess2019IV}.
Recent \acs{lidar}-tailored methods that leverage sensor geometry, re-sample beam layouts~\cite{Kim2023SDG,Langer2020IROS,Ryu2023IDA}, or use 3D surface/occupancy cues~\cite{Yi2021CVPR,Boulch2023Also,Zhang2023ISCC,michele2024saluda}, prove substantially more effective in practice.
These include sensor-aware simulation, surface and occupancy completion from multi-scan aggregation, and geometry-guided label transfer across registered frames~\cite{Sanchez2023ICCV}.
In parallel, representation learning for \acs{lidar} has gained traction.
Occupancy, surface completion, depth auxiliaries, and contrastive objectives help distill transferable 3D features from point clouds.
The resulting encoders can be fine-tuned with fewer labels~\cite{Boulch2023Also,Zhang2023ISCC}.
However, many of these pipelines still rely on some access to the target domain during training, either by means of unlabeled scans for alignment or pseudo-labels for refinement.

While these strategies improve robustness, explicit domain mappings require sensor-specific priors, depend on assumptions about future sensor distributions and may overfit to particular heuristics.
Moreover, they require considerable engineering at training time and their gains can saturate when sensor discrepancies exceed what feature alignment or entropy minimization can bridge.

We address \acs{lidar} \ac{dg} by exploiting geometric structure through an occupancy prediction pretext task during training.
Crucially, unlike prior \ac{dg} work that modifies the input scans (e.g., sensor sparsification, pattern simulation, or domain stylization), our study keeps the input \acs{lidar} data unchanged and alters only the supervision signal.

\figref{fig:title} compares single-scan supervision versus accumulated-points supervision and their effect on an unseen target example side-by-side.
Following~\cite{Boulch2023Also,Rist2021PAMI}, we attach an occupancy decoder and loss to the shared encoder features during training.
At test time, the occupancy head is discarded and we use the encoder–segmentation head to perform semantic segmentation on target scans.
Our training requires only source-domain \acs{lidar} and labels.

We aggregate training targets and evaluate on three large scale \acs{lidar} semantic segmentation datasets SemanticKITTI~\cite{behley2019iccv}, \dnuscenes{}~\cite{fong2021panoptic} and Waymo~\cite{Sun2020CVPR}, which differ in \acs{lidar} models and sampling patterns.
Our approach assumes that source sequences can be registered reliably to generate aggregated supervision.

\noindent{}Summarized, our contributions are:
\begin{itemize}
  \item We propose a geometry-aware auxiliary task that uses accumulated points as supervision during source-domain training.
  \item Our formulation is source-only (no target labels, target sensor models, or target scans), sensor-agnostic, and alters only the supervision signal via an auxiliary head that is discarded at test time.
  \item We evaluate the semantic segmentation task on three large-scale \acs{lidar} datasets covering distinct sensor models and sampling patterns, and provide ablations against standard segmentation training.
\end{itemize}

% !TEX root = ../root.tex
%%

%------------------------------------------------------------------------
\section{Previous Work}

Domain shifts in \acs{lidar} arise both from \emph{sensor-to-sensor} (e.g., beam layout, rotation rate, range noise) and \emph{scene-to-scene} changes (e.g., geography, traffic mix).
Prior work addresses these gaps through generic feature-learning approaches that do not rely on \acs{lidar} specifics, and \acs{lidar}-aware approaches that exploit 3D geometry or measurement physics.
We briefly mention works that use entropy minimization as feature-learning technique used in unsupervised \ac{da}.
While general feature learning is modality-agnostic and broadly applicable, its effectiveness on \acs{lidar} can be limited by sparsity, severe class imbalance, and sensor-induced sampling artifacts.
This motivates the \acs{lidar}-specific strategies reviewed next, which make use of geometric and surface completion, realistic re- and subsampling, and accumulation-based training.
For multi-modal \ac{da} and \ac{dg} we refer to the recent survey~\cite{dong2025mmdasurvey}, and for a \acs{lidar}-focused overview to~\cite{triess2021iv}.

%%
% !TEX root = ../root.tex
%%
\begin{figure*}[t]
  \centering
  % Use the base name without .svg; width spans both columns here
  \includegraphics[width=\textwidth]{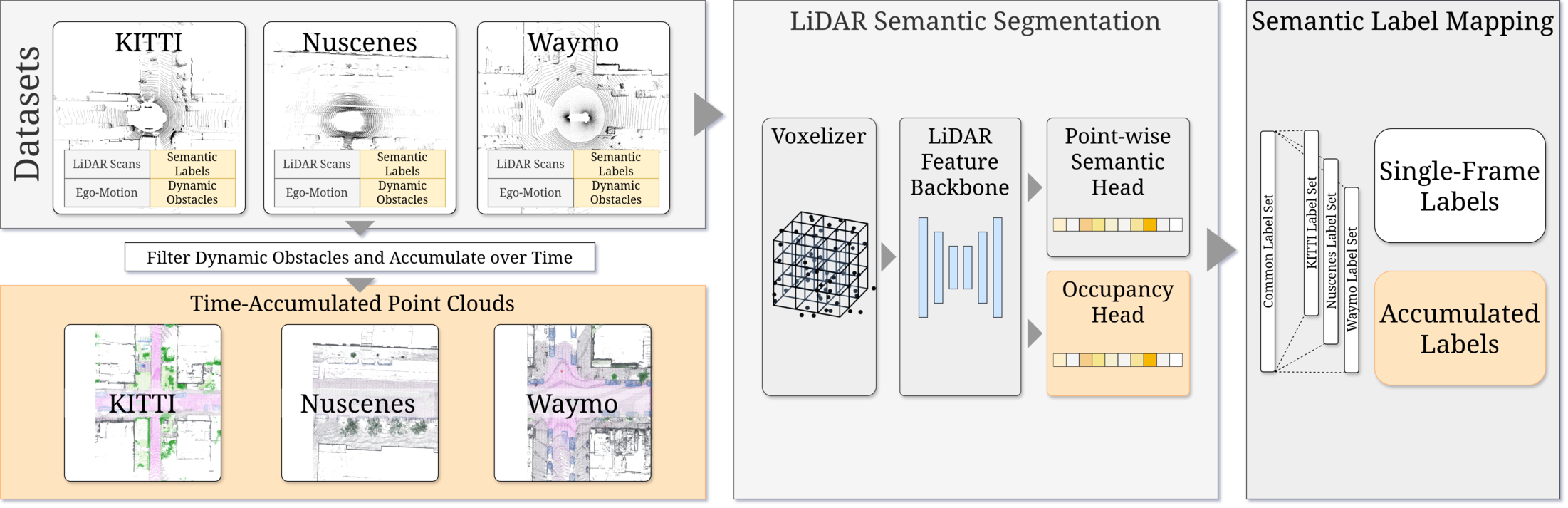} % 
  \caption{%
  \textbf{Proposed method:} %
  We add an occupancy prediction task based on accumulated LiDAR points while training on the source domain to improve model generalization (orange blocks).
  When this task is removed at test time, a vanilla semantic segmentation model remains.
  Even during domain generalization training, the input to the model is unchanged.
  Data for the occupancy prediction is generated automatically on the source domain.
  \label{fig:method}%
  }
\end{figure*}

\subsection{Entropy Minimization}
Entropy minimization is a baseline strategy for \ac{uda}, encouraging confident predictions by minimizing the model's output entropy on unlabeled target data without exploiting sensor- or task-specific structure~\cite{Vu2019CVPR,Chen2019ICCVb,Morerio2018ICLR}.
In dense prediction settings, entropy yields a straightforward objective on the per-point categorical distribution.
Beyond pointwise entropy, aligning the structure of predictions proves beneficial.
Adversarial training in the output space reduces distribution shift by matching target and source prediction layouts rather than only local entropy~\cite{Vu2019CVPR}.
To prevent learning from being dominated by confident target predictions, a maximum-squares loss reshapes the gradient landscape by suppressing gradients for well-classified samples while emphasizing misclassified ones~\cite{Chen2019ICCVb}.
From a statistical-alignment view, Minimal-Entropy-Correlation-Alignment shows that entropy minimization is induced by optimal alignment of second-order statistics between source and target.
Using geodesic instead of Euclidean distances better respects the non-Euclidean curvature of these statistics on symmetric positive-definite manifolds~\cite{Morerio2018ICLR}.

\subsection{Domain Adaptation}

A common pattern in \ac{uda} for images is to attach multi-modal auxiliary tasks that are easy to supervise on both domains, most prominently monocular depth or geometry signals used to guide features and alignment~\cite{Wang2021ICCV,Guizilini2021ICCV,Saha2021CVPR,Wu2022NEURIPS}.
The premise is that learning on complementary, weakly supervised targets produces representations that transfer better to the target domain.
For \acs{lidar}, a pre-training strategy alternating on heterogeneous datasets shows similar robustness benefits~\cite{Rist2019IV}.

Several works exploit 3D geometry to obtain domain-invariant features by predicting surfaces, occupancy, or both~\cite{Yi2021CVPR,Boulch2023Also,Zhang2023ISCC,michele2024saluda}.
Complete~\!\&\!~Label~\cite{Yi2021CVPR} first completes sparse voxels to a canonical surface representation and then segments it.
Because the intermediate 3D proxy is independent of a sensor’s sampling pattern, the approach bridges \sensortosensor{} gaps and outperforms generic entropy-based \ac{da}.
Follow-ups predict occupancy (and sometimes semantics) in the neighborhood of a query point and use such heads as pre-training to improve segmentation and detection with limited target labels~\cite{Boulch2023Also,Zhang2023ISCC}.
SALUDA~\cite{michele2024saluda} replaces explicit completion with an implicit surface head jointly trained with segmentation so that source and target share a common surface representation.
This yields gains on real-to-real and sim-to-real benchmarks.
More broadly, occupancy prediction has also been shown to be an effective proxy task for learning transferable 3D representations.
SPOT~\cite{yan2025spot} pre-trains LiDAR backbones via semantic occupancy prediction and beam re-sampling augmentation, yielding robust cross-dataset transfer across downstream detection and segmentation tasks under varying sensor configurations.

Another line performs explicit \sensortosensor{} mapping by resampling the training data.
This can operate on single scans or on temporally accumulated point clouds~\cite{Kim2023SDG,Langer2020IROS,Ryu2023IDA}.
Instead of predicting a canonical surface as in~\cite{Yi2021CVPR}, these methods produce semi-synthetic scans with target-like properties.
To assess the fidelity of such artificial scans, Triess~\etal{} propose a learned realism metric for \acs{lidar} point clouds, showing that local point cloud realism correlates with downstream segmentation performance~\cite{Triess2022IJCV}.
Surface-aware re-simulation can be computationally heavy when explicit recovery is needed~\cite{Langer2020IROS}, while z-buffer ray casting on accumulations is faster~\cite{Ryu2023IDA}.
A second stage can align statistics between domains, for example by geodesic correlation alignment to generalize beyond a single \sensortosensor{} pair toward \datasettodataset{} shifts~\cite{Morerio2018ICLR,Wu2019ICRA}.

Finally, mixing samples from different sensors and training a student with pseudo-labels from a teacher improves target performance without requiring target annotations~\cite{Saltori2022Cosmix}.

\subsection{Domain Generalization}

Kim~\etal{}~\cite{Kim2023SDG} propose \emph{DGLSS}, which trains on a single source and simulates unseen sensors by random subsampling with the main objective to generalize from denser to sparser sensor domains.
Two density-aware constraints align features between dense and sparsified views and stabilize class relations.
This label-free strategy increases robustness but requires a sufficiently dense source sensor.
The inverse direction from sparse to dense sensor domains is harder to achieve with DGLSS:
The augment baseline and the SCC-only variant outperform the full method in this setting.
Sánchez~\etal{}~\cite{Sanchez2023ICCV} aggregate multiple sweeps and propagate labels to accumulated points, improving robustness across sensors.
Their method includes an explicit geometric association step that assigns labels to accumulated points using hand-designed rules.
The resulting labeled points are then used both as supervision and as (subsampled) model input.
In contrast, our approach introduces a free-space class directly into the accumulation process and trains jointly on occupied and free-space points, without modifying the input representation.

Kim~\etal{}~\cite{Kim2024ECCV} target generalization by explicitly learning across diverse densities within the source domain.
A density-aware feature module and targeted density augmentation strengthen robustness of \acs{lidar} segmentation.
Ryu~\etal{}~\cite{Ryu2023CVPR} synthesize target-like frames from accumulated source clouds using detailed sensor models that account for ego-motion, spin distortions, and sampling patterns.
This removes the need for target data but requires accurate sensor specifications to apply realistic ego-motion and spin distortions as well as sampling patterns.
Zhao~\etal{}~\cite{Zhao2024BMVC} define a \ac{dg} setting for synthetic-to-real indoor segmentation and improve robustness with point-level pattern augmentations and instance mixing.

\subsection{Enriched Supervision Signal}
All of the above \ac{dg} methods operate exclusively on source data but alter the source-domain method inputs via subsampling, accumulation, mixing, or sensor-aware rendering to expose the model to domain diversity without any target samples.
There are notable works in image classification/segmentation showing that enriching the supervision signal alone can improve robustness.
Incorporating self-supervised tasks (jigsaw puzzle~\cite{Carlucci2019CVPR}, rotation prediction~\cite{Qin2025IET}), predicting depth along with labels in multi-task learning~\cite{Ren2018CVPR}, or privileged simulation depth as an alternative label modality~\cite{Vu2019ICCV} have led to more domain-invariant features.
These works demonstrate that a model's generalization can be significantly affected by richer supervision, all while keeping the input data unchanged.

% !TEX root = ../root.tex
%%

%------------------------------------------------------------------------

\section{Proposed Method}

\subsection{Overview}
We aim to train a \acs{lidar} semantic segmentation network that generalizes to unseen sensor domains without altering the input scan at test time.
To this end, we introduce an additional \emph{occupancy prediction} pretext task during training, see \figref{fig:method}.
The point-cloud semantic segmentation problem is defined over a dataset $\cD=\{(P_i, y_i, p_i)\}_{i=1}^{N}$ of \acs{lidar} sweeps $P_i$, point-wise labels $y_i$, and ego poses $p_i$. The model predicts $y_i$ from $P_i$.

Using the ego-motion, we aggregate successive sweeps into a denser, static world frame and sample \emph{free-space} points along each registered \acs{lidar} ray between the sensor origin and the first return.
This yields an enriched supervision signal with more labeled surface points via temporal accumulation, and additional free-space points that supervise scene completion.
At test time on both source and unseen target domains, the occupancy branch is simply ignored and the network is used as a standard point-cloud semantic segmenter.

% !TEX root = ../root.tex
%%

\newcommand{\overlayimg}[5][]{%
  \begin{tikzpicture}
    % Pick width (default \linewidth)
    \def\owidth{\ifx\relax#1\relax\linewidth\else#1\fi}
    % Bottom image
    \node[inner sep=0, opacity=#4] (base) {\includegraphics[width=\owidth]{#2}};
    % Keep bbox = image
    \path[use as bounding box] (base.south west) rectangle (base.north east);
    % Optional: comment the next 'scope' if you don't want blend behavior grouped
    \begin{scope}[transparency group] % , blend mode=multiply % (uncomment to multiply)
      % Top image
      \node[inner sep=0, opacity=#5] at (base.center) {\includegraphics[width=\owidth]{#3}};
    \end{scope}
  \end{tikzpicture}%
}

\newcommand{\labeledoverlayW}[5]{%
  \begin{tikzpicture}[x=1pt,y=1pt]
    \pgfmathsetlengthmacro{\cropW}{\linewidth}
    \pgfmathsetlengthmacro{\cropH}{(2/3)*\linewidth}
    % Reserve some space on the left for the vertical label (tweak as needed)
    \pgfmathsetlengthmacro{\labelpad}{16pt}
    
    \path[use as bounding box] (-.5*\cropW,-.5*\cropH) rectangle (.5*\cropW,.5*\cropH);
    \begin{scope}
      \clip (-.5*\cropW-\labelpad,-.5*\cropH) rectangle (.5*\cropW,.5*\cropH);
      % Scale to match crop WIDTH (so tall/narrow sources "cover"; excess height is clipped)
      \node[inner sep=0, opacity=#4] at (.5*\labelpad,0)
        {\includegraphics[width=\cropW,keepaspectratio]{#2}};
      \begin{scope}[transparency group] % , blend mode=multiply
        \node[inner sep=0, opacity=#5] at (.5*\labelpad,0)
          {\includegraphics[width=\cropW,keepaspectratio]{#3}};
      \end{scope}
    \end{scope}
    \node[rotate=90, anchor=center] at (-.5*\cropW+.5*\labelpad,0) {\small\bfseries #1};
  \end{tikzpicture}%
}

\begin{figure*}[t]
  \centering
  % Column (a)
  \resizebox{\linewidth}{!}{%
  \begin{subfigure}[t]{0.32\textwidth}
    \centering
    \labeledoverlayW{\dkitti{}}{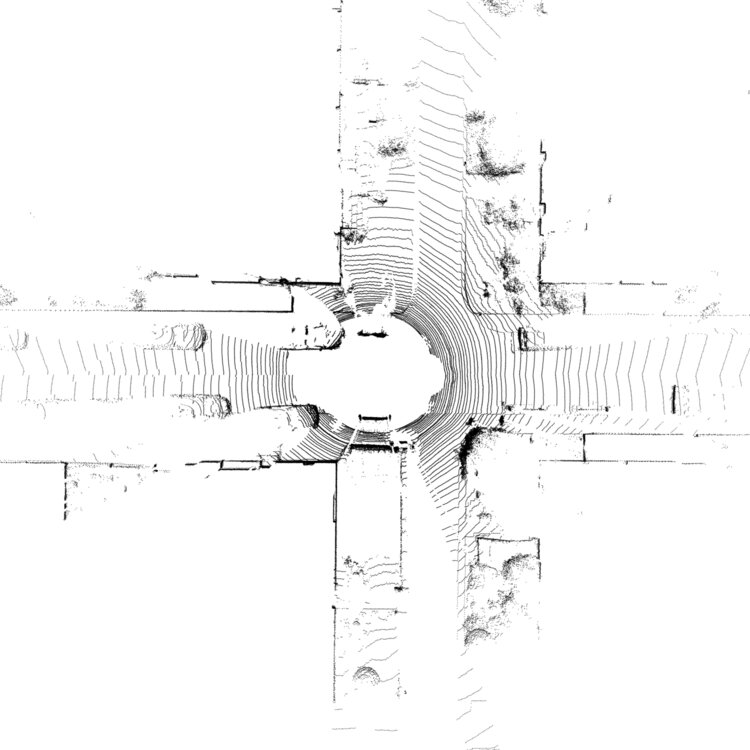}{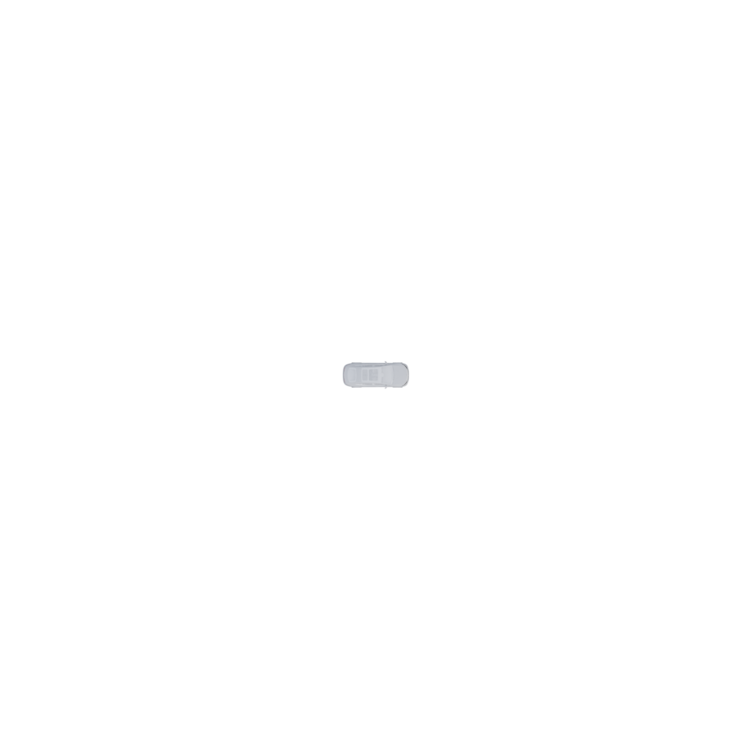}{1.0}{1.0}\par\vspace{4pt}
    \labeledoverlayW{\dnuscenes{}}{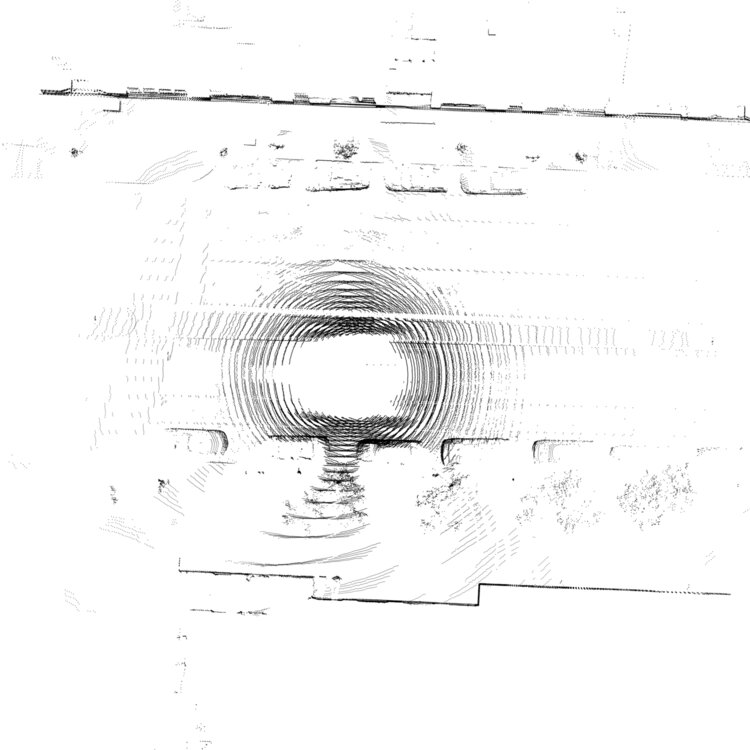}{vehicle_only_camera_top.png}{1.0}{1.0}\par\vspace{4pt}
    \labeledoverlayW{\dwaymo{}}{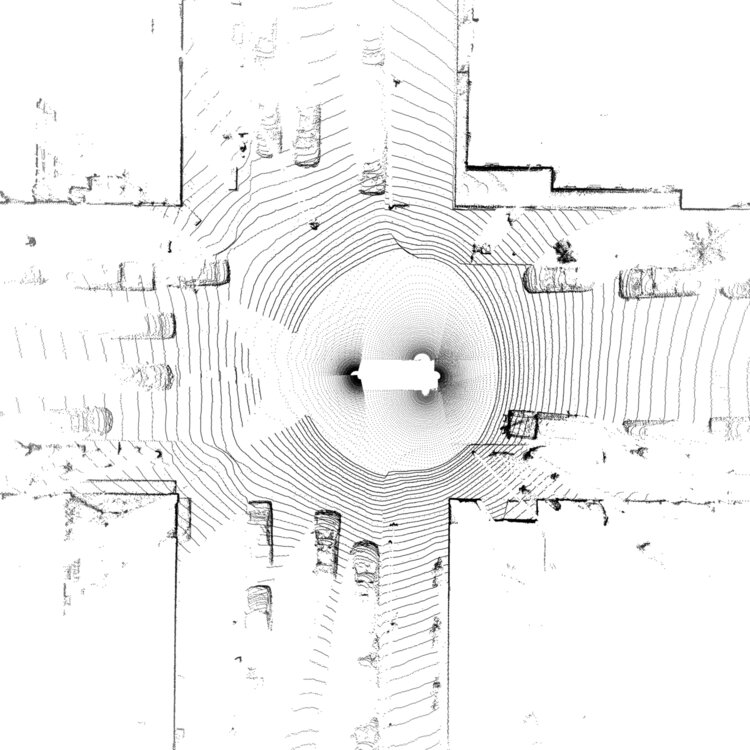}{vehicle_only_camera_top.png}{1.0}{1.0}
    \subcaption{LiDAR scan bird's-eye view}
  \end{subfigure}\hfill
  \hspace{2mm}
  % Column (b)
  \begin{subfigure}[t]{0.32\textwidth}
    \centering
    \overlayimg{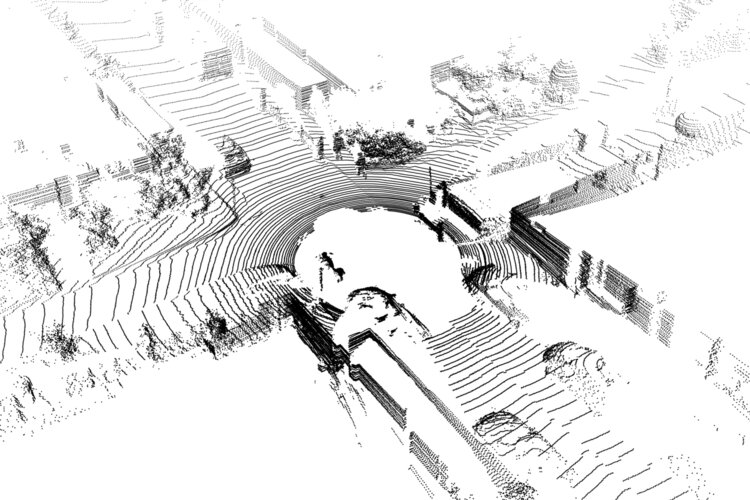}{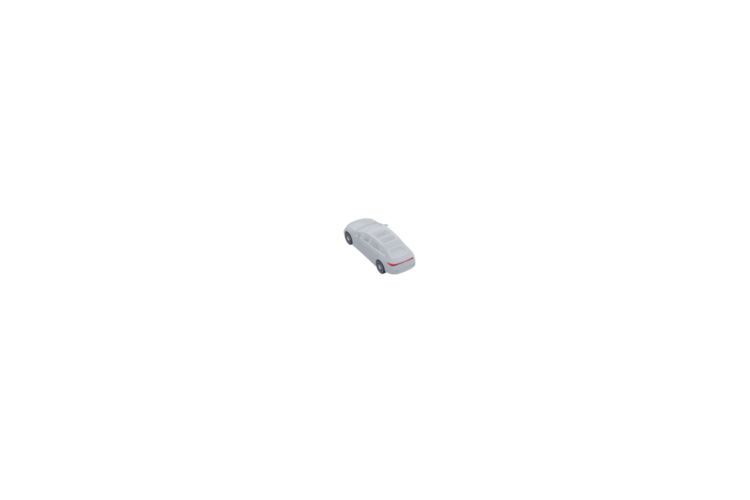}{1.0}{1.0}\par\vspace{4pt}
    \overlayimg{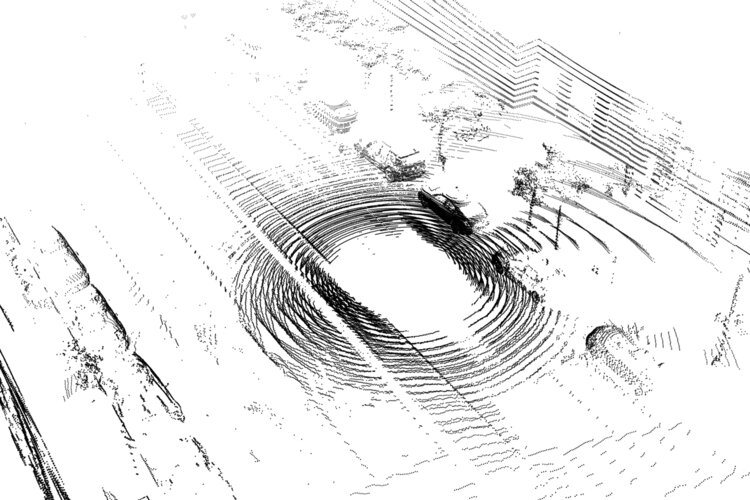}{vehicle_only_camera_main.png}{1.0}{1.0}\par\vspace{4pt}
    \overlayimg{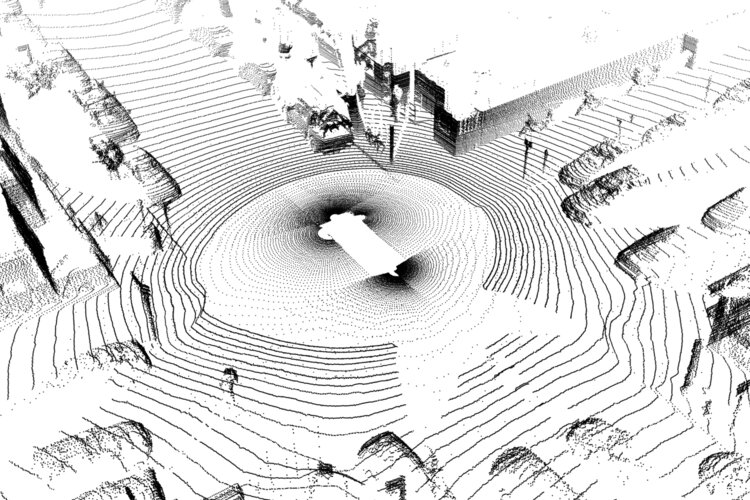}{vehicle_only_camera_main.png}{1.0}{1.0}
    \subcaption{LiDAR scan perspective view}
  \end{subfigure}\hfill
  \hspace{2mm}
  % Column (c)
  \begin{subfigure}[t]{0.32\textwidth}
    \centering
    \includegraphics[width=\linewidth]{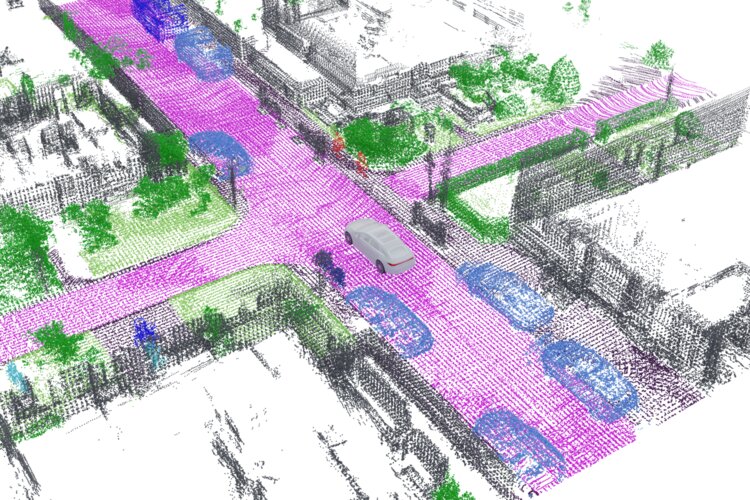}\par\vspace{4pt}
    \includegraphics[width=\linewidth]{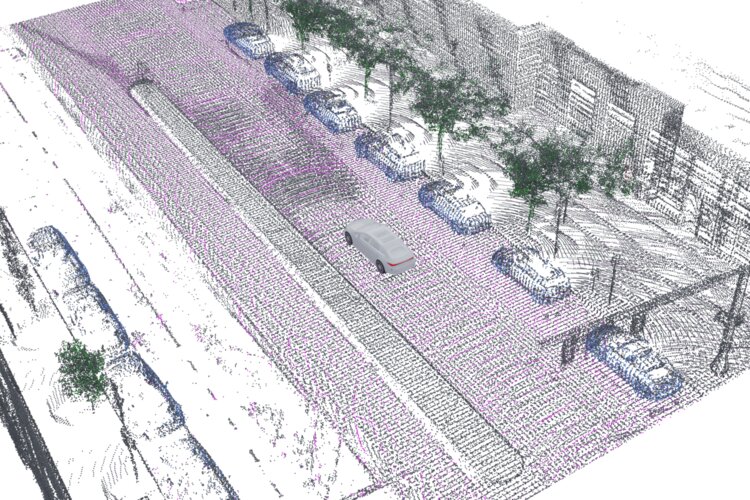}\par\vspace{4pt}
    \includegraphics[width=\linewidth]{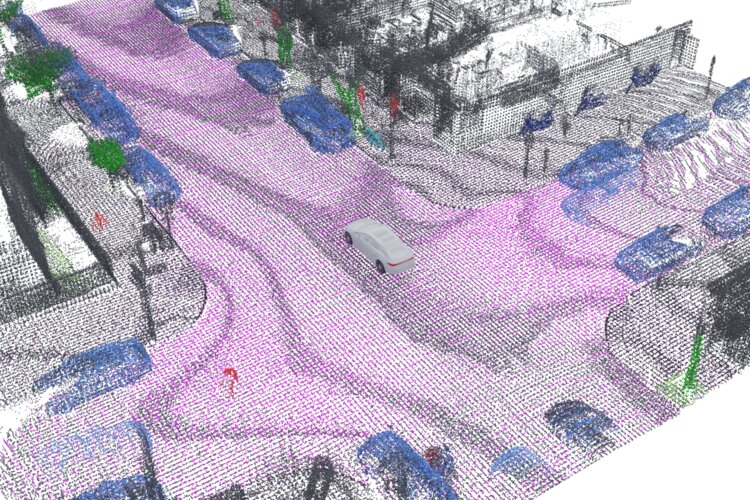}
    \subcaption{World-frame accumulated points}
  \end{subfigure}
  } % resizebox
  \caption{\textbf{Dataset characteristics.} Accumulated world-frame ground truth is semantically color-coded by the \emph{10-A} label set. Rows illustrate differing \acs{lidar} configurations across datasets. KITTI uses a single 64-beam \acs{lidar}, yielding up to 115k points within 100 ms. nuScenes fuses the five most recent 32-beam scans over 250 ms. Waymo employs multiple \acsp{lidar} to reduce blind spots.}
  \label{fig:dataset-comp}
\end{figure*}

\subsection{Training Data}
We construct the occupancy supervision by \emph{accumulating \acs{lidar} scans over time} in a common world frame using the provided ego-motion.
For \dnuscenes{}, we refine the provided poses with an ICP alignment over local scan windows before fusion, improving registration accuracy.
We retain an individual \acs{lidar} point in the accumulation only if its distance to its own recording sensor origin is below \SI{100}{\meter}.
This cap filters far-range, typically noisier returns and reduces the resulting point density.
To keep training efficient, we limit the number of retained points per voxel, which is particularly important near the ego vehicle where raw point density is highest.
After aggregation, we place a 3D voxel grid with an edge length of \SI{16}{\centi\meter} over the dense point cloud to thin out high-density regions.
When downsampling within a voxel, we preferentially retain labeled points to maximize the density of semantic supervision.
This higher density of labeled surface points and reduced overlap among points describing the same geometry improves training efficiency as each retained point carries more supervision signal.
This procedure defines our canonical, training domain shared across datasets.
Visually, the right column of \figref{fig:dataset-comp} shows that the accumulated point clouds exhibit similar spatial structure across \dkitti{}, \dnuscenes{}, and \dwaymo{}.

% Handling dynamic obstacles
To avoid painting motion trails into the fused world model, we explicitly remove scan points belonging to dynamic objects from all non-current frames.
We identify dynamic objects via either a dedicated dynamic flag (\dkitti{}) or via 3D annotated bounding boxes with temporal interpolation to the \acs{lidar} timestamp (\nuscenes{}, \dwaymo).
Concretely, each non-current sweep is time-aligned to the reference frame, its boxes are interpolated to the sweep time, and all points falling inside any dynamic box are excluded from accumulation.
Only the dynamic-object points observed at the \emph{current} ego frame are retained in the resulting training example, preserving instantaneous geometry without leaving history traces in the static world model.

%Free-space supervision
During training, we generate \emph{free-space} supervision by sampling points along registered \acs{lidar} rays between the sensor and the first return, labeling them as non-occupied.
The model thus learns semantic segmentation on a larger set of surface points from temporal fusion, and a scene-completion signal via one extra “free-space” class on sampled ray points.

We use only raw 3D coordinates (no reflectivity), following \cite{Sanchez2023ICCV,Kim2024ECCV}.
We do not sub-sample or augment the source-domain sweeps based on domain assumptions.
Only raw 3D points are consumed by the pre-processing and the model.
\tabref{tab:dataset_info} lists the training split sizes (scenes and annotated points, in millions) for \dkitti{}, \dnuscenes{}, and \dwaymo{}, including pairwise and joint combinations.
As expected, temporal accumulation increases the number of semantically labeled points compared to single sweeps.

\subsection{Model}
Our network (\figref{fig:method}) takes a raw 3D point list, voxelizes it into point pillars, and processes them with a multi-resolution backbone~\cite{Lang2019CVPR}.
Predictions are produced by a Coordinate MLP head that can be queried at arbitrary 3D locations~\cite{Rist2021PAMI}.
This design makes it straightforward to train either only on the input \acs{lidar} scan in a default semantic segmentation setup or on the accumulated surface points for semantic segmentation and sampled free-space points for occupancy, without changing the output head.
In practice, we query the head at the accumulated \acs{lidar} point coordinates and at free-space sample locations.
At test time we simply drop the free-space output class, normalize probabilities for only the semantic classes and query only the \acs{lidar} points.
Alternatively, the accumulated-point training can be implemented on a voxel grid by pooling point labels into voxels and supervising a scene-completion grid.
Conversely, standard backprojection (“unpooling”) used in (sparse)-voxel and BEV-conv architectures can map grid outputs back onto the original surface and free-space samples to provide an equivalent supervision signal.

%------------------------------------------------------------------------
\section{Experiments}

\subsection{Existing Approaches and Label Sets}
We compare against three representative methods:
C\&L (Yi~\etal{}~\cite{Yi2021CVPR}) is an unsupervised domain adaptation approach that leverages access to unlabeled target-domain data.
DGLSS (Kim~\etal{}~\cite{Kim2023SDG}) is a domain generalization method that simulates unseen sensors by subsampling the input scan.
3DLabelProp (Sánchez~\etal~\cite{Sanchez2023ICCV}) also accumulates points, but uses them both as supervision and (subsampled) model input.
Both \cite{Yi2021CVPR} and \cite{Kim2023SDG} employ MinkowskiNet~\cite{Choy2019Minkowski}-based architectures and deliver source-domain segmentation performance comparable to ours.

We consider the widely used \kitti{}, \nuscenes{}, and \dwaymo{} \acs{lidar} segmentation datasets.
All three are automotive datasets with large numbers of outdoor \acs{lidar} scans captured in sequences and annotated with point-wise semantics.
Because each dataset defines its own taxonomy, we map classes to two variants of a 10-class intersection label set.
Variant \emph{10-A}, introduced by Yi~\etal{}\cite{Yi2021CVPR}, contains:
car, bicycle, motorcycle, truck, other vehicle, person, drivable surface, sidewalk, terrain, vegetation, infrastructure.
Variant \emph{10-B} from Kim~\etal{}\cite{Kim2023SDG} merges \emph{terrain} and \emph{vegetation} and distinguishes \emph{sidewalk} from \emph{walkable}, yielding a slightly different 10-class mapping used in semantic DG evaluation.

\subsection{Dataset Characteristics}
\begingroup

\sisetup{
  parse-numbers = true,
  table-number-alignment = center,
  round-mode = places,
  round-precision = 0,   % one decimal place
  group-separator = {,},
  group-minimum-digits = 4
}

\newlength{\knwcola}
\setlength{\knwcola}{9.5mm} %
\newcolumntype{K}{>{\centering\arraybackslash}p{\knwcola}}

\begin{table}[htb]
    \noindent %
    \centering
    \setlength{\tabcolsep}{2pt}
    \caption{Number of \acs{lidar} semantic segmentation labels in the dataset training splits, including all pairwise combinations (table rows). %
    Training samples ("scenes") and number of annotated \acs{lidar} points are distinguished (table columns).}
    \label{tab:dataset_info}

    \resizebox{\linewidth}{!}{%
\begin{tabular}{@{}KKK
  S[table-format=6.0, table-number-alignment=center]
  S[table-format=5.0, table-number-alignment=center]
  S[table-format=5.0, table-number-alignment=center]
@{}}
  \toprule
        & & &
        \multicolumn{1}{c}{\multirow{2}{*}{\shortstack[c]{\\\textbf{Samples}}}} &
        \multicolumn{1}{c}{\multirow{2}{*}{\shortstack[c]{\textbf{Semantic}\\\textbf{Labels}}}} &
        \multicolumn{1}{c}{\multirow{2}{*}{\shortstack[c]{\textbf{Accumulated}\\\textbf{Labels}}}} \\
        \multicolumn{3}{c}{\cellcolor[HTML]{a9c4fc}\textbf{Dataset $\downarrow$}} & & \\
        
        \cmidrule(lr){5-6}
        KITTI & NuScenes & Waymo & & \multicolumn{2}{c}{in Million Points} \\
        \midrule
        \checkmark{} & & & 19310 & 1776.439484 & 2684.298945 \\
        & \checkmark{} & & 27430 & 518.611106  & 1832.199564 \\
         & &\checkmark{} & 23691 & 2479.763464 & 3671.013059 \\
%        %
       \arrayrulecolor{gray!60}\specialrule{0.4pt}{1pt}{1pt}\arrayrulecolor{black}
       \checkmark{}  &\checkmark{}  & & 46740 & 2295.050590 & 4516.498509 \\
       \checkmark{}  & &\checkmark{}  & 43001 & 4256.202948 & 6355.312004 \\
         &\checkmark{}  &\checkmark{}  & 51121 & 2998.374570 & 5503.212623 \\
%        %
       \arrayrulecolor{gray!60}\specialrule{0.4pt}{1pt}{1pt}\arrayrulecolor{black}
       \checkmark{}  &\checkmark{}  &\checkmark{}  & 70431 & 4774.814054 & 8187.511568\\
        \bottomrule
    \end{tabular}
    }
\end{table}

\endgroup

\begingroup

\sisetup{
  parse-numbers = true,
  table-number-alignment = right,
  round-mode = places,
  round-precision = 1,
}

\newlength{\knwcol}
\setlength{\knwcol}{9.5mm} %
\newcolumntype{K}{>{\centering\arraybackslash}p{\knwcol}}
\newcolumntype{N}{S[table-format=2.1, table-number-alignment=right, table-column-width=\knwcol]}

\newcommand{\hlS}[1]{\cellcolor{gray!15}\tablenum{#1}}
\newcommand{\hlA}[1]{\cellcolor{green!15}\tablenum{#1}}

\begin{table*}[htbp]
  \setlength{\tabcolsep}{3.5pt}
  \centering
  \caption{\textbf{Domain generalization performance (10-B)} of our proposed approach (left, center) and DGLSS~\cite{Kim2023SDG} (right).
  Values are mIoU scores of the semantic segmentation task on the 10-B label set (higher is better).
  A gray background highlights source-domain evaluation, green background highlights improvements on the unseen target domain with accumulated points training over the segmentation baseline.
  }
  \label{tab:dataset_confusion_10b}
  \resizebox{\textwidth}{!}{%
   
  \begin{tabular}{K K K N N N N N N N N N}
    \toprule
    & & &
    \multicolumn{3}{c}{Acc. Points Training} &
    \multicolumn{3}{c}{Baseline Segmentation} &
    \multicolumn{3}{c}{DGLSS~\cite{Kim2023SDG}} \\
    \multicolumn{3}{c}{\cellcolor[HTML]{a9c4fc}\textbf{Trained on $\downarrow$}} &
    \multicolumn{9}{c}{\cellcolor[HTML]{FFCC99}\textbf{Evaluated on $\downarrow$}} \\
    \cmidrule(lr){4-6}\cmidrule(lr){7-9}\cmidrule(lr){10-12}
    {\dkitti{}} & {\dnuscenes{}} & {\dwaymo{}} &
    {\dkitti{}} & {\dnuscenes{}} & {\dwaymo{}} &
    {\dkitti{}} & {\dnuscenes{}} & {\dwaymo{}} &
    {\dkitti{}} & {\dnuscenes{}} & {\dwaymo{}} \\
    \midrule
    \checkmark &          &               & \hlS{58.29} & 34.53 & 35.98 & \hlS{61.05} & 36.39 & 36.33 & \hlS{59.62} & 44.83 & 40.67 \\
    & \checkmark &                        & 47.22 & \hlS{70.85} & \hlA{52.54} & 48.72 & \hlS{69.71} & 44.98 & {-} & {-} & {-} \\
    &  & \checkmark                       & 47.47 & \hlA{50.24} & \hlS{73.74} & 49.16 & 43.41 & \hlS{74.70} & 51.23 & 49.61 & \hlS{75.28} \\
    \arrayrulecolor{gray!60}\specialrule{0.4pt}{1pt}{1pt}\arrayrulecolor{black}
    \checkmark & \checkmark &             & \hlS{58.16} & \hlS{68.20} & \hlA{51.98} & \hlS{62.87} & \hlS{68.84} & 50.60 & {-}    & {-}    & {-}    \\
    \checkmark & &       \checkmark       & \hlS{60.65} & \hlA{54.86} & \hlS{73.01} & \hlS{61.57} & 52.68 & \hlS{72.22} & {-}    & {-}    & {-}    \\
    & \checkmark & \checkmark             & 51.68 & \hlS{66.42} & \hlS{71.14} & 52.04 & \hlS{68.44} & \hlS{72.48} & {-}    & {-}    & {-}    \\
    \arrayrulecolor{gray!60}\specialrule{0.4pt}{1pt}{1pt}\arrayrulecolor{black}
    \checkmark & \checkmark & \checkmark  & \hlS{57.94} & \hlS{61.69} & \hlS{64.37} & \hlS{63.56} & \hlS{65.38} & \hlS{68.87} & {-}    & {-}    & {-}    \\
    \bottomrule
  \end{tabular}

 } 
\end{table*}

\endgroup

\begingroup

\newlength{\knwcolx}
\setlength{\knwcolx}{14.5mm}
\newcolumntype{K}{>{\centering\arraybackslash}p{\knwcolx}}
\newcolumntype{N}{S[table-format=2.2, table-number-alignment=right, table-column-width=\knwcolx]}

\newcommand{\hlS}[1]{\cellcolor{gray!15}\tablenum{#1}}
\newcommand{\hlA}[1]{\cellcolor{green!15}\tablenum{#1}}

\begin{table*}[t]
  \setlength{\tabcolsep}{1.7pt}
  \centering
  \caption{\textbf{Domain generalization performance (10-A)} of our scene completion approach (left), our baseline semantic segmentation (middle) and previous work (right).
  Semantic segmentation on 10-A label set. Values are mIoU scores. Higher is better.
  A gray background highlights source-domain evaluation.
  }
  \label{tab:dataset_confusion_10a}
    
  \resizebox{\linewidth}{!}{%
   
  \begin{tabular}{K K N N N N N N N N}
    \toprule
    & &
    \multicolumn{2}{c}{Acc. Points Training} &
    \multicolumn{2}{c}{Baseline Segmentation} &
    \multicolumn{2}{c}{C\&L~\cite{Yi2021CVPR}} &
    \multicolumn{2}{c}{3DLabelProp~\cite{Sanchez2023ICCV}} \\
    \multicolumn{2}{c}{\cellcolor[HTML]{a9c4fc}\textbf{Trained on $\downarrow$}} &
    \multicolumn{8}{c}{\cellcolor[HTML]{FFCC99}\textbf{Evaluated on $\downarrow$}} \\
    \cmidrule(lr){3-4}\cmidrule(lr){5-6}\cmidrule(lr){7-8}\cmidrule(lr){9-10}
    {\dkitti{}} & {\dnuscenes{}} &
    {\dkitti{}} & {\dnuscenes{}} &
    {\dkitti{}} & {\dnuscenes{}} &
    {\dkitti{}} & {\dnuscenes{}} &
    {\dkitti{}} & {\dnuscenes{}} \\
    \midrule
    \checkmark &             & \hlS{64.8} & \hlA{32.5} & \hlS{62.1} & 28.3 & \hlS{50.2} & 31.6 & \hlS{69.0} & 42.7 \\
    & \checkmark             & \hlA{46.2} & \hlS{67.7} & 45.2 & \hlS{71.2} & 33.7 & \hlS{54.4} & 50.5 & \hlS{66.5} \\
    \bottomrule
  \end{tabular}

 } 
\end{table*}

\endgroup

The \acs{lidar} configurations differ substantially.
\dkitti{} is recorded with a 64-beam Velodyne HDL-64 at \SI{10}{\hertz};
\dnuscenes{} uses a 32-beam Velodyne VLP-32 at \SI{20}{\hertz};
the Waymo \acs{lidar} segmentation dataset employs multiple proprietary roof-mounted \acsp{lidar} in a coordinated setup.
Beyond beam count and spin rate, vertical beam elevation patterns and multi-sensor layouts yield distinct sampling geometries and point densities, visible in our scan comparison (Fig.~\ref{fig:dataset-comp}).
On \nuscenes{}, we follow common practice for its sparser frames and use the last five \acs{lidar} scans as input.
\tabref{tab:dataset_info} lists the numbers of training samples, semantic labels, and accumulated point labels per dataset and dataset combination.
We use the default training/validation splits of each dataset.
Accumulation increases the number of labeled \acs{lidar} points by roughly \SI{50}{\percent} on \kitti{} and \dwaymo{}, and by about \(\sim\!3.5\times\) on \nuscenes{}.

\subsection{Single-Source Domain Generalization}

% 10-B comparison
The single-source DG results are reported in the top sections of \tabref{tab:dataset_confusion_10b} and \tabref{tab:dataset_confusion_10a}.
All values are mean Intersection-over-Union (mIoU, \%) for semantic segmentation on the respective validation splits.

\paragraph{Results on \emph{10-B}.}
\tabref{tab:dataset_confusion_10b} compares our source-only segmentation baseline, our proposed accumulated-points (scene-completion) training, and DGLSS~\cite{Kim2023SDG}.
Source-domain performance is nearly unchanged between segmentation and accumulated-points training; only on \dkitti{} we observe a slight drop with accumulated points.
DGLSS attains slightly higher source-domain scores.
On the target domain, trends depend on the dataset pair:
Training on \dnuscenes{} or \dwaymo{} yields substantial improvements toward the other dataset with our accumulated-points supervision (\SI{+8}{\percent} and \SI{+7}{\percent}, highlighted in green).
In contrast, generalization to or from \dkitti{} slightly degrades relative to the segmentation baseline.
Accordingly, our method modestly outperforms DGLSS for \dwaymo{},$\rightarrow$,\dnuscenes{} transfers, but not for transfers involving \dkitti{}.
\figref{fig:showcase} illustrates four qualitative examples, showing improved robustness to scan-pattern artifacts on ground surfaces and crisper object boundaries.

\paragraph{Results on \emph{10-A}.}
The \emph{10-A} results are similar to the above.
Transfers involving \dkitti{} benefit somewhat more than under \emph{10-B}, with accumulated-points supervision improving robustness over the segmentation baseline (marked in green).
From \dkitti{} to \dnuscenes{}, accumulated-points training can surpass C\&L’s domain adaptation relative to the segmentation baseline.
3DLabelProp achieves higher domain generalization performance overall, indicating that altering the input point cloud remains a strong approach for cross-domain robustness.

\subsection{Ablations on Single-Source and Multi-Source Domain Generalization}

We observe that source-domain performance is very similar with and without the occupancy (scene-completion) auxiliary task.
The bottom block of \tabref{tab:dataset_confusion_10b} reports multi-source training.
All three pairwise combinations (\dkitti+\dnuscenes, \dkitti+\dwaymo, \dnuscenes+\dwaymo, rows 4-6) and the all-sources setting (\dkitti+\dnuscenes+\dwaymo, row 7).
Training on all three datasets simultaneously slightly reduces per-dataset source accuracy but still exceeds the single-source domain generalization results reported above for both the segmentation baseline and the accumulated-points variant.
Adding \dkitti{} to the training mix benefits the segmentation baseline the most, nearly closing the gap to its best single-source domain generalization performance.
However, accumulated-points training on \dkitti+\dnuscenes{} and \dkitti+\dwaymo{} continues to outperform the corresponding segmentation baselines.
Relative to single-source training, gains are modest for \dkitti+\dwaymo{} (row 5) and largely neutral for \dkitti+\dnuscenes{} (row 4).
For transfer to \dkitti{}, combining \dnuscenes{} and \dwaymo{} (row 6) provides a clear boost over single-source training—interestingly, in this two-source combination the segmentation baseline attains the better score.

\begingroup

\setlength{\tabcolsep}{2pt}
\newlength{\colw}
\setlength{\colw}{\dimexpr(\textwidth - 8\tabcolsep)/5\relax} 
\newcolumntype{C}{>{\centering\arraybackslash}m{\colw}}

\newcommand{\img}{\includegraphics[width=\linewidth,height=22mm,keepaspectratio]{example-image}}
\newcommand{\imgfile}[1]{\includegraphics[width=\linewidth,height=22mm,keepaspectratio]{#1.jpg}}

\newcommand{\tabsection}[2]{%
  % \addlinespace[-3pt]
  \multicolumn{#1}{@{}c@{}}{\footnotesize\scshape #2}\\[-4pt]
  \arrayrulecolor{gray!60}\cmidrule(lr){1-#1}\arrayrulecolor{black}
}

% \multicolumn{3}{c}{\cellcolor[HTML]{a9c4fc}\textbf{Trained on $\downarrow$}} &
% \multicolumn{9}{c}{\cellcolor[HTML]{FFCC99}\textbf{Evaluated on $\downarrow$}} \\

\begin{figure*}[t]
  \centering
  \renewcommand{\arraystretch}{1.0}
  \begin{NiceTabular}{@{} C C C C C @{}}
\CodeBefore
      \begin{tikzpicture}
        \node[
          draw=gray,
          line width=1.2pt,
          rounded corners=1pt,
          fit=(2-|2) (4-|4),
          inner sep=-2pt
        ] {};
        \node[
          draw=gray,
          line width=1.2pt,
          rounded corners=1pt,
          fit=(5-|4) (7-|6),
          inner sep=-2pt
        ] {};
        \node[
          draw=green!60,
          line width=1.2pt,
          rounded corners=1pt,
          fit=(2-|4) (4-|5),
          inner sep=-2pt
        ] {};
        \node[
          draw=green!60,
          line width=1.2pt,
          rounded corners=1pt,
          fit=(5-|2) (7-|3),
          inner sep=-2pt
        ] {};
      \end{tikzpicture}
    \Body
\tabsection{5}{Evaluation on nuScenes Dataset}
\imgfile{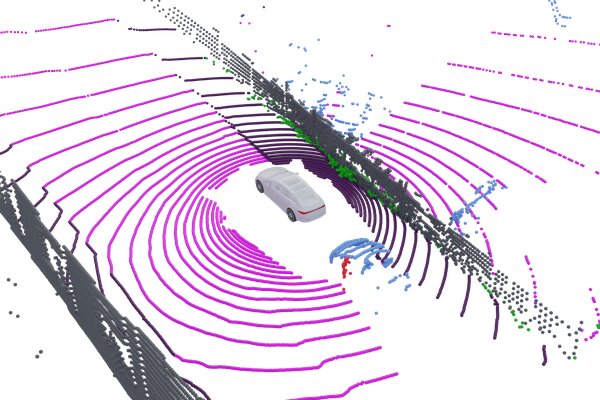} & %
\imgfile{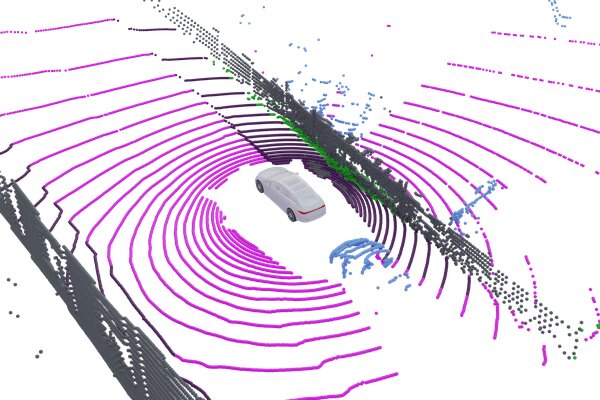} & %
\imgfile{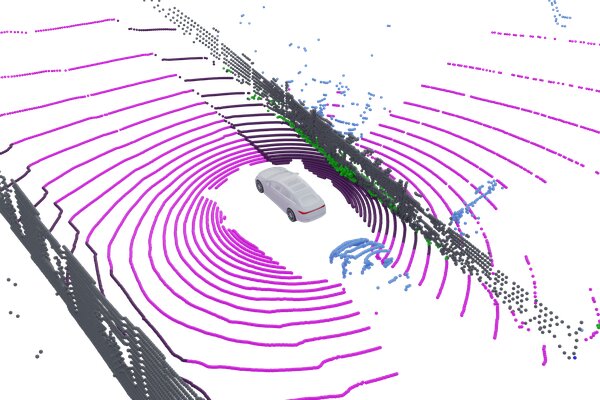} & %
\imgfile{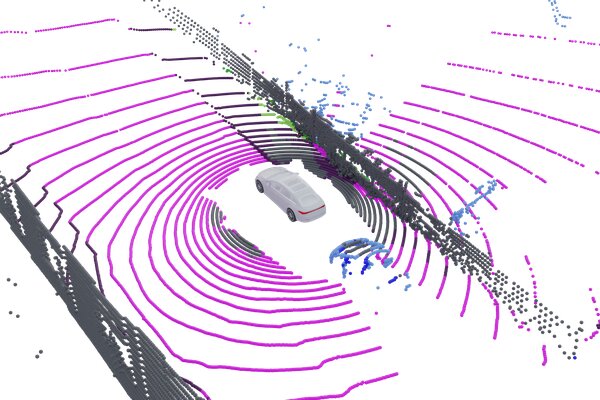} & %
\imgfile{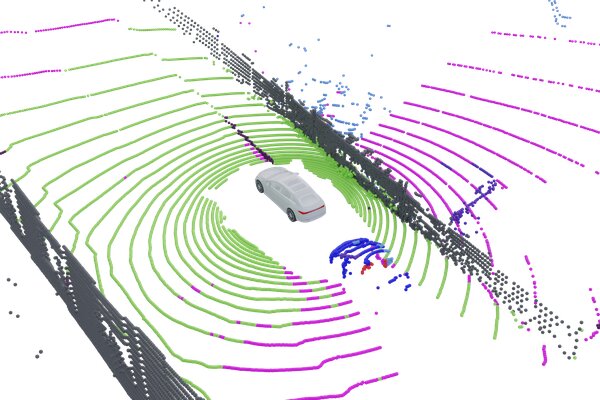} \\ %
\imgfile{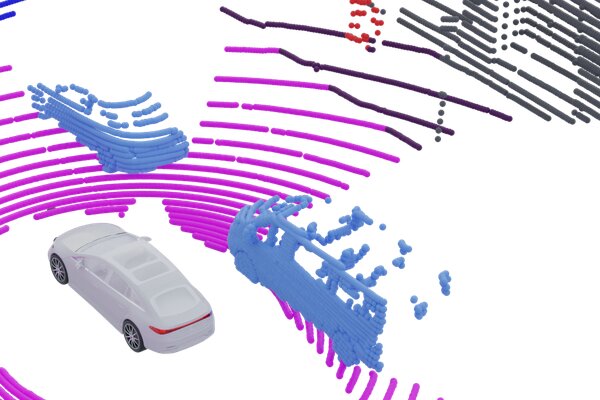} & %
\imgfile{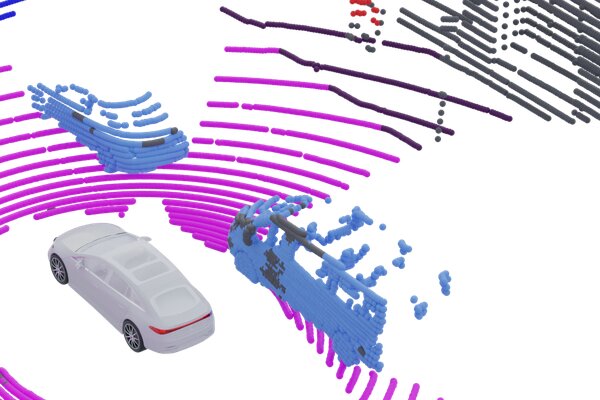} & %
\imgfile{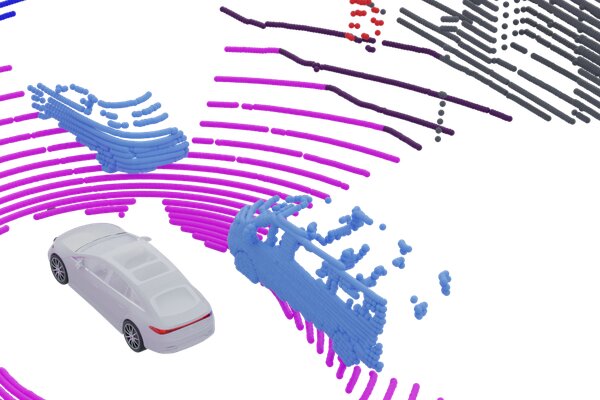} & %
\imgfile{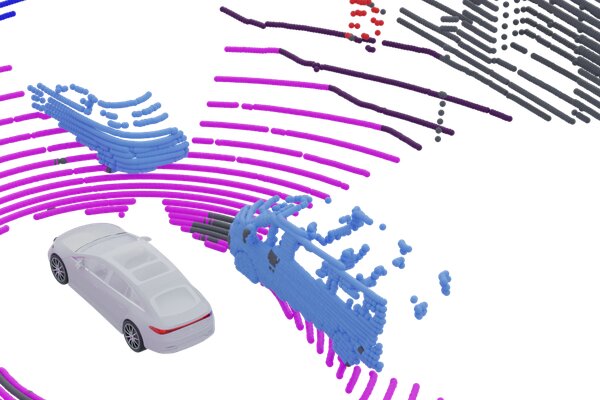} & %
\imgfile{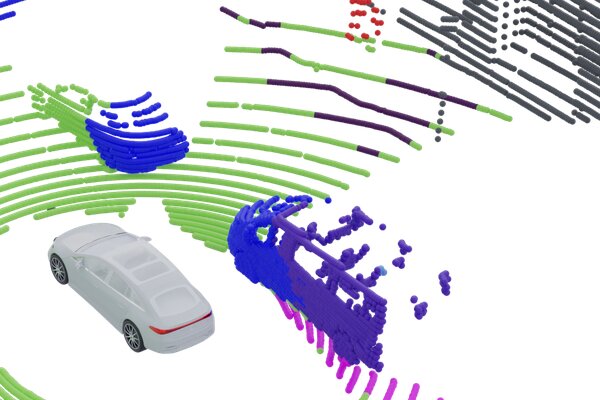} \\ %
\tabsection{5}{Evaluation on Waymo Dataset}
\imgfile{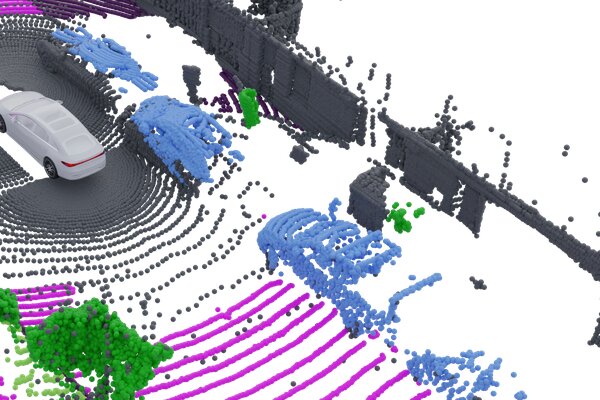} & %
\imgfile{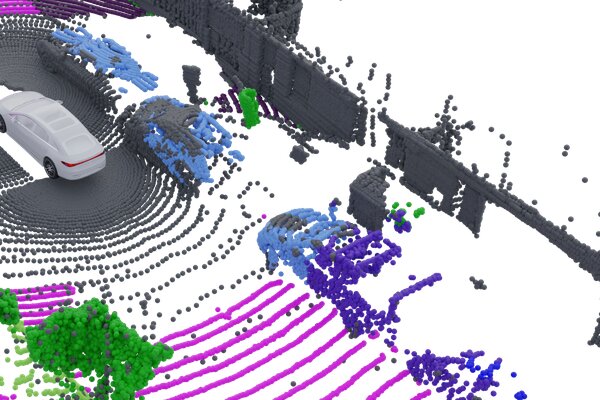} & %
\imgfile{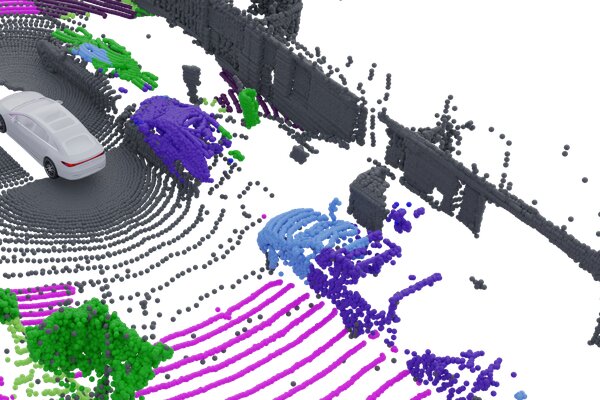} & %
\imgfile{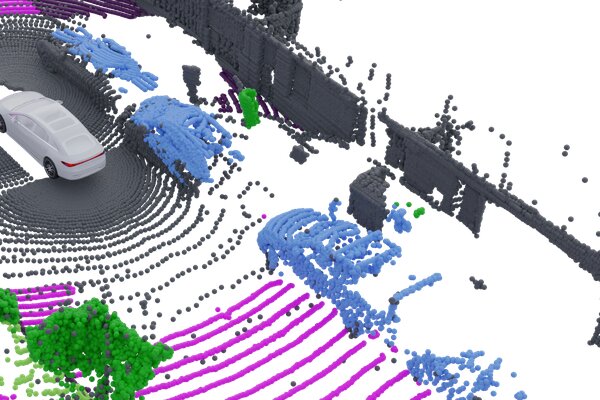} & %
\imgfile{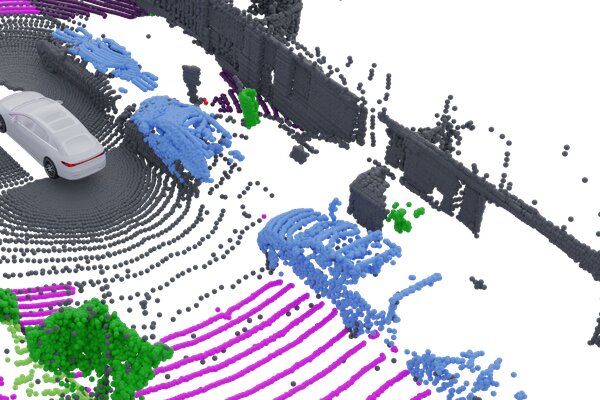} \\ %
\imgfile{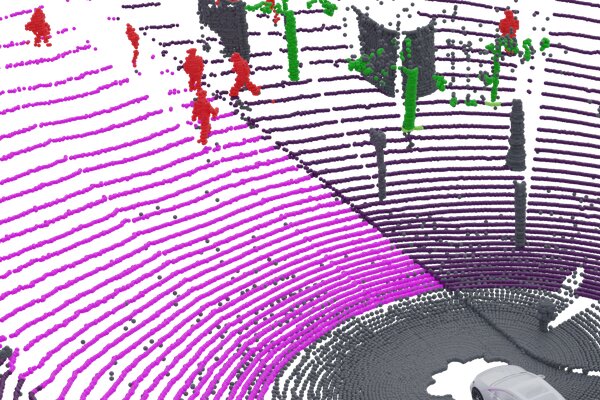} & %
\imgfile{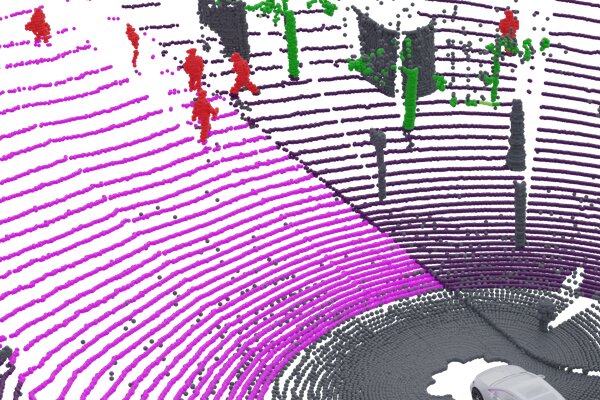} & %
\imgfile{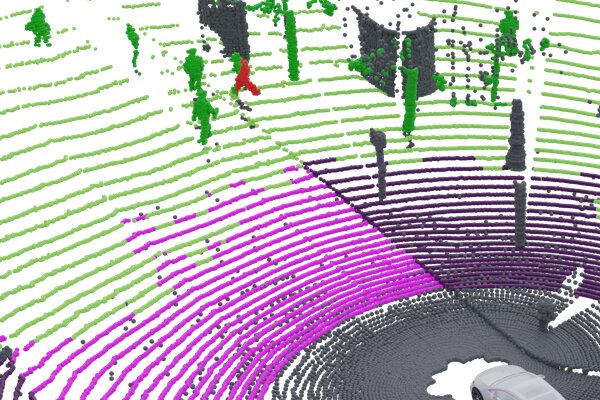} & %
\imgfile{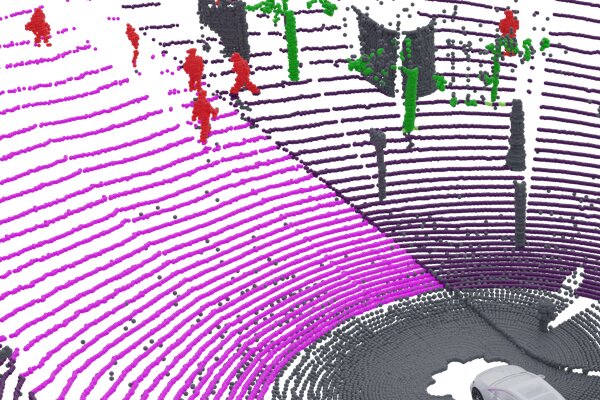} & %
\imgfile{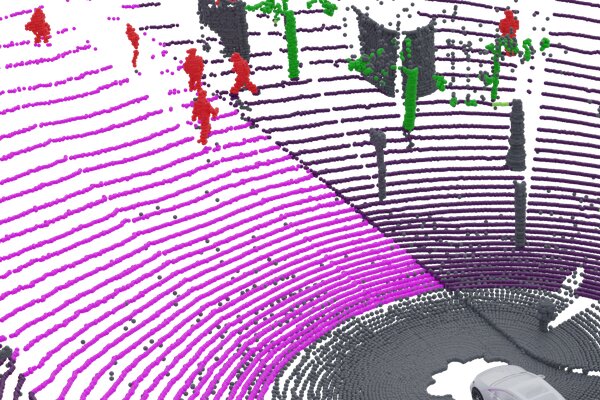} \\ %
    \subcaption*{Ground Truth} &
    \subcaption*{(a) Acc. Points nuScenes} &
    \subcaption*{(b) Sem Segm. nuScenes} &
    \subcaption*{(c) Acc. Points Waymo} &
    \subcaption*{(d) Sem Segm. Waymo} \\[-6pt]
\end{NiceTabular}
  \caption{\textbf{Qualitative Results:} %
  Examples of cross-sensor domain generalization with two evaluation examples from the nuScenes and two from the Waymo dataset (one per row).
  Each row displays, from left to right: ground truth reference, segmentation result trained on (a) nuScenes accumulated points, (b) nuScenes segmentation, (c) Waymo accumulated points, (d) Waymo segmentation.
  For clarity, all unlabeled points (gray) are shown as unlabeled in the prediction columns.
  A gray border highlights source-domain evaluation; green background highlights improvements on the unseen target domain.
  Training on accumulated multi-scan points in the source domain yields a noticeably more robust network.
  Ground segmentation issues due to scan patterns are effectively reduced and object boundaries are more consistent.
  Importantly, no visible regression is observed on the source domain when using accumulated points compared to standard segmentation.
  }
  \label{fig:showcase}
\end{figure*}

\endgroup

\subsection{Discussion}

Our results indicate that enriching the supervision signal with accumulated-points (scene-completion) targets yields a domain-robustness benefit at almost no cost to source-domain accuracy.
The observed source trade-off is negligible, indicating that the added objective does not overfit or dilute features critical for in-domain segmentation.
This points less to a model-capacity limitation and more to supervision shaping, meaning that completion targets encourage geometry- and boundary-aware features that generalize well across sampling patterns.
Importantly, our method does not alter the input point cloud at any stage.
It isolates the effect of richer supervision from the effect of input transformations seen in prior DG/DA work, clarifying that more informative outputs from the same inputs can improve feature robustness.

The effectiveness nonetheless depends on the source–target pairing.
Gains are pronounced for \dwaymo{}\,$\leftrightarrow$\,\dnuscenes{}, while transfers involving \dkitti{} remain more difficult.
We do not attribute this specifically to scan pattern or density as these factors likely differ sufficiently across all datasets.
Instead, broader dataset-to-dataset domain gaps might be more influential.
\dnuscenes{} and \dwaymo{} are captured primarily in dense urban environments, whereas \dkitti{} spans a mix of less densely populated areas, highways, and urban scenes.
In particular, the \dkitti{} validation split offers limited diversity, which may skew cross-domain results. Future work should verify this with targeted experiments on additional datasets.
Differences in scene composition and geographic context, object frequency and scale, and the balance between static and dynamic classes can all reduce the utility of features learned in one domain when applied to another.

Multi-source training further indicates that a simple data pooling is no solution to the domain generalization problem.
Adding \dkitti{} can narrow gaps for source-only baselines, and accumulated-points supervision generally improves performance but with modest gains. 
It might be necessary to address the domain factors in a more targeted way.
Overall, requesting more outputs from the same network on unchanged scans is a simple, stable lever for domain generalization.
It is effective when inter-domain semantics overlap sufficiently.
Beyond that, richer geometric supervision on unchanged scans forces the network to learn mid-level, geometry-aware cues, which improves robustness to domain shift.
This is aligned with prior image-domain results where adding geometric supervision forces models to learn more comprehensive representations and transferable high-level semantics.
We observe the same effect for \acs{lidar}.
%

% !TEX root = ../root.tex
%%

%------------------------------------------------------------------------
\section{Conclusion}

We approached LiDAR domain generalization by keeping the input scan unchanged while enriching supervision with a scene-completion task from accumulated points, and we evaluated this setup on combinations of \dkitti{}, \dnuscenes{}, and \dwaymo{}.
Our experiments show that switching to a richer supervision signal, while keeping the input scan unchanged, can considerably shift domain-generalization outcomes for \acs{lidar} segmentation.
To our knowledge, prior work has not isolated this effect without modifying the input itself.
We observe strong dataset-pair dependence.
Transfers involving KITTI remain challenging, whereas \dwaymo{}-\dnuscenes{} generalize substantially better.
Although our approach does not surpass the very latest domain generalization methods, it is competitive with a classic surface-based \ac{da} baseline and with a more recent \ac{dg} approach.
Our approach is complementary to previous work.
Input-preserving supervision can be combined with input-altering strategies to aggregate their strengths.

Next, we propose to adapt our setup to \ac{uda} by deriving target-domain completion labels and using them only as auxiliary supervision during training.
This adds supervision while keeping single-scan inference unchanged, testing whether target-side completion can close the remaining gap.

% !TEX root = ../root.tex
%%

% {\small
% \bibliographystyle{template/ieee_fullname}
% %\bibliography{ConvAbrvLong,template/egbib,IEEEabrv,references}
% \bibliography{ConvAbrvLong,IEEEabrv,references}
% }

%\addtolength{\textheight}{-12cm}
%\vspace{10mm}
\bibliographystyle{IEEEtran}
% Your .bib file here
\bibliography{ConvAbrvLong,IEEEabrv,references}

\end{document}